\documentclass[lettersize,journal]{IEEEtran}
\usepackage{amsmath,amsfonts}
\usepackage{array}
\usepackage[caption=false,font=normalsize,labelfont=sf,textfont=sf]{subfig}
\usepackage{textcomp}
\usepackage{stfloats}
\usepackage{url}
\usepackage{verbatim}
\usepackage{graphicx}
\usepackage{algpseudocode}
\usepackage{subfig}
\usepackage{booktabs}
\usepackage{multirow}
\usepackage{algorithm}

\def\BibTeX{{\rm B\kern-.05em{\sc i\kern-.025em b}\kern-.08em
    T\kern-.1667em\lower.7ex\hbox{E}\kern-.125emX}}
\usepackage{balance}

\begin{document}
\title{RA-Nav$:$ A Risk-Aware Navigation System Based on Semantic Segmentation for Aerial Robots in  Unpredictable Environments}
\author{Ziyi Zong, Xin Dong, Jinwu Xiang, Daochun Li, Zhan Tu}



\maketitle

\begin{abstract}
Existing aerial robot navigation systems typically plan paths around static and dynamic obstacles, but fail to adapt when a static obstacle suddenly moves. Integrating environmental semantic awareness enables estimation of potential risks posed by suddenly moving obstacles.
In this paper, we propose RA-Nav, a risk-aware navigation framework based on semantic segmentation. A lightweight multi-scale semantic segmentation network identifies obstacle categories in real time. These obstacles are further classified into three types: stationary, temporarily static, and dynamic. For each type, corresponding risk estimation functions are designed to enable real-time risk prediction, based on which a complete local risk map is constructed. Based on this map, the risk-informed path search algorithm is designed to guarantee planning that balances path efficiency and safety. Trajectory optimization is then applied to generate trajectories that are safe, smooth, and dynamically feasible. Comparative simulations demonstrate that RA-Nav achieves higher success rates than baselines in sudden obstacle state transition scenarios. Its effectiveness is further validated in simulations using real-world data.
\end{abstract}


\section{Introduction}
Aerial robots (e.g., drones) are highly maneuverable, fast, and low cost, making them increasingly popular in aerial photography, 3D scene reconstruction, and narrow environment exploration\cite{c3,c4}. However, in these applications, aerial robots encounter complex obstacles with variable states. In particular, static pedestrians may suddenly start moving, or pedestrians may unexpectedly emerge from behind buildings. Such sudden state transitions of obstacles present significant challenges to autonomous navigation and potential risk prediction.

To navigate safely in unpredictable real-world environments, aerial robots need to not only perceive and understand their surroundings, but also detect potentially mobile objects and assess risks from occluded or unknown areas. This requires proactive planning with awareness of risk to avoid dangerous situations in advance.

However, many methods primarily focus on structural representations of the environment\cite{c4,c5,c6}, considering only the positions and shapes of visible static and dynamic obstacles. They lack the capability to infer whether a seemingly static obstacle might become dynamic. Recent studies have explored the use of semantic information to enhance autonomous navigation. Some of these methods focus on autonomous driving, emphasizing road semantics \cite{c7}. Nevertheless, most approaches primarily address static objects while neglecting hidden or uncertain environmental risks \cite{c8}. End-to-end learning-based approaches offer promising alternatives but often require extensive training data and powerful computing platforms\cite{c9,c10}. Furthermore, all these methods are designed for navigation in either continuously dynamic or purely static environments, neglecting risk estimation for suddenly moving obstacles in real-world environments.

\begin{figure*}
    \centering
    \includegraphics[width=1.0\linewidth]{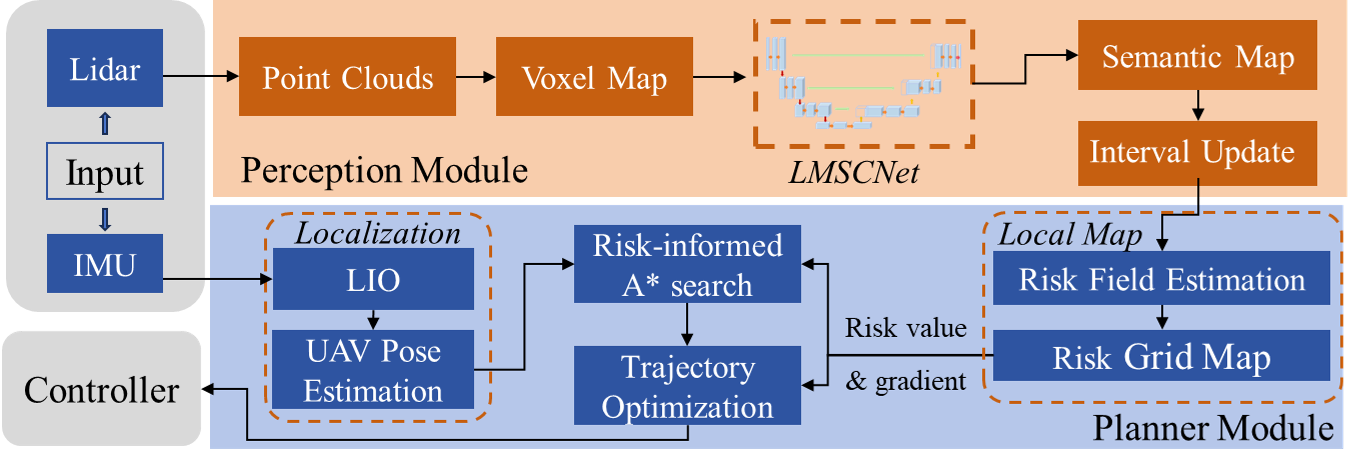}
    \caption{Framework of the semantic-enhanced navigation system. The perception and planning modules run asynchronously on the onboard computer. The perception module uses U-Net for semantic segmentation and updates a local map at fixed intervals. The planning module then uses the risk grid map for safe path search and optimization.}
    \label{framework}
\end{figure*}

To enhance the autonomous navigation capability of aerial robots in unpredictable real-world environments, this work introduces a risk-aware navigation framework based on semantic segmentation, as shown in Fig.\ref{framework}. The proposed framework consists of two main modules. The first one is a lightweight semantic completion network. The network processes sparse point clouds to quickly and accurately infer obstacle semantics. These semantic results are then integrated into a local map for risk assessment and path planning.
The second one is the planning module, in which we design category-specific risk estimation functions for different types of obstacles. For static obstacles, a two-dimensional Gaussian distribution is applied, leading to risk reduction with increasing distance. For dynamic obstacles, we propose a velocity-dependent risk model to capture their anisotropic and time-varying motion characteristics. By combining the static and dynamic risk fields, a unified environmental risk map is generated, encoding both spatial and temporal risk information.
Based on this risk map, we propose a novel risk-informed path search method balancing path efficiency and safety. Furthermore, a risk-based objective function is formulated to optimize the trajectory, allowing the aerial robot to quickly replan in response to sudden obstacle motion and ensuring safe and continuous navigation.

In summary, the main contributions of this work are as follows.

\begin{enumerate}
    \item A semantic-based risk grid map is constructed to provide fine-grained environmental understanding.
    \item A risk-informed path search strategy is proposed to account for both efficiency and safety.
    \item A semantic-aware planning framework is proposed to ensure safe navigation in high-risk scenarios involving unpredictable obstacle motion and occlusions.
\end{enumerate}   

The remainder of the paper is organized as follows. Section \ref{related works} reviews related work on semantic segmentation and path planning. In section \ref{LBSCNet}, we introduce the semantic segmentation network. In section \ref{model}, we present the risk models for different categories of obstacles. In section \ref{planning}, we introduce the risk-aware path planning method in detail. Section \ref{simulation} presents the simulated  results. Finally, we summarize our work in Section \ref{conclusions}.

\section{Related works}
\label{related works}
In recent years, semantic map-based path planning has recently attracted significant attention in aerial robot navigation. A major challenge remains how to incorporate environmental semantics into dynamic planning algorithms to enhance robustness in complex and high-risk scenarios.
\subsection{Semantic Segmentation
}
Semantic map construction is a key component of environmental understanding and is typically categorized into vision-based and LiDAR-based approaches. Vision sensors are widely used due to their low cost and rich semantic content. Chen et al.\cite{c11} leveraged the DeepLab-v3+ architecture to perform pixel-wise semantic segmentation, enabling dense semantic mapping with category labels for Aerial Robot navigation. Xiong et al. \cite{c12} combined RGB-D cameras with BiSeNetV1 to perform 3D semantic reconstruction in agricultural environments, although efficiency in dense scenes remains a concern. Hu et al. \cite{c13} employed an improved RandLA-Net for indoor semantic mapping and used KNN to enhance object localization accuracy, but its generalization in complex environments is limited. Roemer et al. \cite{c14} integrated uncertainty-aware probabilistic segmentation to propose a safety-aware obstacle avoidance strategy in dynamic environments, yet their method remains dependent on visual sensors, which limits the accuracy of depth perception under challenging conditions.

LiDAR sensors, with their high-precision point cloud capabilities, offer stronger adaptability in dynamic settings. Miller et al. \cite{c15} introduced an air-ground collaborative system that shares LiDAR-based semantic maps to improve multi-robot coordination. Chen et al. \cite{c16} developed the RangeSeg framework for real-time semantic segmentation and 3D map construction to support planning. Li et al. \cite{c17} proposed SD-SLAM, which integrates semantic information and Kalman filtering to distinguish between dynamic and static landmarks, improving mapping accuracy in dynamic scenes. Xie et al. \cite{c18} and Li et al. \cite{c19} each proposed lightweight LiDAR semantic segmentation methods—based on CNNs and center-focused networks respectively—providing feasible solutions for low-power Aerial Robot platforms. To cope with rapidly changing environments, researchers have also focused on improving response times. Ding et al. \cite{c20} introduced LENet, balancing real-time performance and lightweight design, although robustness in complex scenarios remains to be improved.  Roldao et al. propose LMSCNet\cite{c21}, a new multi-scale 3D semantic scene completion method, and its lightweight feature makes it suitable for mobile robot navigation. Lightweight 3D semantic occupancy networks, such as SCONet\cite{c29}, exhibit low prediction accuracy and produce incomplete local maps in cluttered environments. Although using transformers \cite{c30} or 3D CNNs \cite{c31} can improve accuracy, they are not suitable for resource-constrained aerial robots.
\subsection{Path Planning}
In terms of planning, classic algorithms like RRT \cite{c22} address nonholonomic constraints with asymptotic optimality, while graph-based methods such as A* \cite{c23} provide efficient solutions on discretized grids with guaranteed completeness, and its dynamic variant D* \cite{c24} further enables replanning in changing environments, making them widely adopted in robotic navigation. However, these geometric methods are unable to distinguish semantically different areas (e.g., stationary obstacles vs. temporarily static obstacles), often leading to suboptimal decisions. To address this, Galindo et al. \cite{c25} and Kantaros et al. \cite{c26} incorporated semantic maps into task planning to enhance autonomy, but their approaches struggle with adaptation in dynamic environments. Qian et al. \cite{c27} proposed an adaptive navigation framework that combines semantic SLAM with online probabilistic planning, achieving improved flexibility in unknown environments, though at the cost of high computational complexity. Hu et al. \cite{c13} constructed a semantic map using RandLA-Net and proposed a semantic-aware A* algorithm, which significantly improves safety and planning efficiency. However, its generalization in complex indoor environments remains to be validated.

In summary, existing studies remain constrained by three key issues: the robustness of semantic mapping under highly dynamic or densely populated conditions is insufficient; lightweight algorithms often compromise accuracy, making it difficult to balance real-time performance and precision; and aerial robot path planning approaches require better adaptability to occlusions and sudden obstacle emergence. To address these challenges, this work proposes a semantic map-based planning framework that integrates dynamic semantic map updates with adaptive trajectory optimization, enhancing aerial robot navigation performance in high-risk, real-world environments.

\section{Lightweight Multi-scale Semantic Segmentation}
\label{LBSCNet}
In this section, we present the semantic segmentation network (LMSCNet \cite{c21}) designed for sparse point cloud data. The network follows a U-Net architecture, consisting of a 2D convolutional backbone and a 3D segmentation head, and employs upsampling modules to perform multi-scale 3D semantic segmentation and completion with low computational complexity. By deploying a pretrained model offline on our onboard device, LMSCNet enables real-time environment segmentation and predicts the distribution of occluded obstacles. Based on these predictions, a risk field is constructed to support subsequent path planning.
\subsection{Semantic Segmentation Network Architecture}
\begin{figure}
    \centering
    \includegraphics[width=1.0\linewidth]{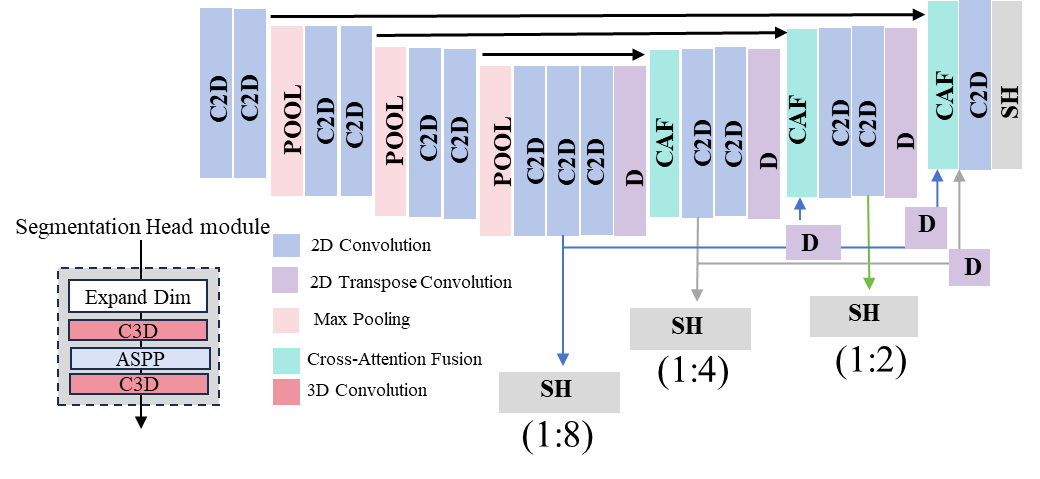}
    \caption{Semantic segmentation network architecture, composed of a 2D backbone and a 3D semantic segmentation head, which enables multi-scale semantic segmentation.}
    \label{Net}
\end{figure}
The network takes sparse 3D voxel grids as input and produces dense semantic voxel labels through an encoder–decoder architecture. To infer dense scene representations from sparse inputs, we employ a standard 4-layer U-Net structure, where each layer reduces the resolution by a factor of two via convolution and pooling operations, thereby progressively enlarging the receptive field.

In the encoder, LMSCNet adopts 2D convolutions to maintain a lightweight design while ensuring efficiency in handling sparse inputs. In particular, the convolutional operations are restricted to the X–Y plane, with the height dimension (Z) projected onto the feature channels, which enables an effective compression of the voxel representation. Multi-scale skip connections enable feature propagation across different resolutions, facilitating contextual modeling and providing fine-grained semantic cues for decoding.

The decoder integrates 3D convolutions to progressively recover the full 3D scene structure. During upsampling, high-resolution spatial features are fused with low-resolution global context features, enhancing the reconstruction of details and complex structures. Unlike approaches that project point clouds into 2D bird’s-eye views, LMSCNet directly processes the 3D voxel space, avoiding structural losses caused by dimensionality reduction.

For the semantic segmentation head, the network employs a combination of dense and dilated convolutions, followed by an Atrous Spatial Pyramid Pooling (ASPP) module to achieve multi-scale feature fusion. By increasing dilation rates, ASPP aggregates contextual information from multiple receptive fields. However, as dilated convolutions are less suited for sparse inputs, we further integrate dense 3D convolutions within the ASPP to densify the feature representation, mitigate sparsity effects, and preserve consistency with backbone features.

To improve multi-scale feature fusion, we introduce a cross-attention module at each skip connection. In this module, high-resolution features serve as the query, while low-resolution features are used as key and value. A multi-head attention mechanism performs weighted fusion, with the query and key dimensions reduced via 1×1 convolutions to reduce computational cost. The fused features are then combined with the original high-resolution features using residual connections to preserve semantic integrity.

To meet the real-time requirements of aerial robots in dynamic environments, LMSCNet adopts a multi-scale prediction strategy. At lower resolutions, the network provides coarse scene completion for global planning and safety assessment, while higher resolutions offer fine-grained semantic segmentation for local obstacle avoidance and detailed understanding. This multi-scale inference mechanism enhances the flexibility of scene understanding while reducing computational load and memory usage.

During training, we apply a cross-entropy loss at each scale \cite{c33}, ensuring that predictions across all resolutions are jointly optimized. When the model identifies previously unknown object categories, subsequent path planning can adopt a more cautious strategy, thereby improving safety and robustness in complex and uncertain environments.

\section{Semantics-based Obstacle Risk Assessment}
\label{model}
In complex real-world environments, obstacles vary in characteristics, and their dynamics critically affect aerial robot navigation safety. To address this, we classify obstacles into three types and design corresponding risk models for risk-aware planning:

Type I: Static obstacles rigidly attached to the ground (e.g., buildings, walls, trees), whose risks are low.

Type II: Continuously dynamic obstacles (e.g., vehicles, pedestrians) with predictable motion patterns, whose risks depend on their velocities.

Type III: Temporarily static but potentially dynamic obstacles (e.g., standing pedestrians, parked vehicles), which are often misclassified as static but may suddenly move, posing the highest collision risk.

In the following section, we detail the modeling of different obstacles and their role in constructing risk fields, whose values and gradients later guide path search.
\subsection{Risk Model for Static Obstacles}

\begin{figure}[!t]
\centering
\subfloat[\label{fig:a1}]{
		\includegraphics[scale=0.30]{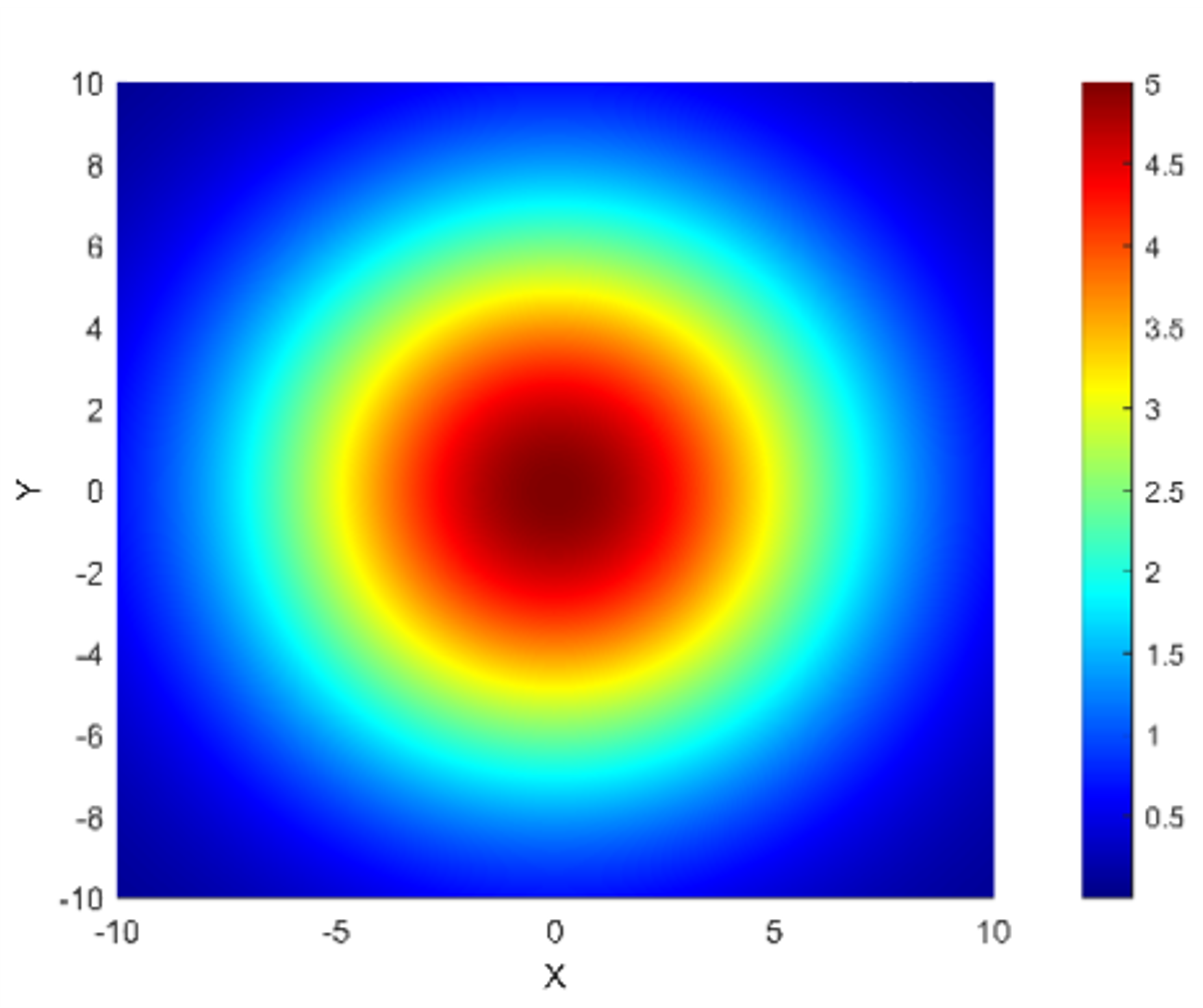}}
\subfloat[\label{fig:b1}]{
		\includegraphics[scale=0.30]{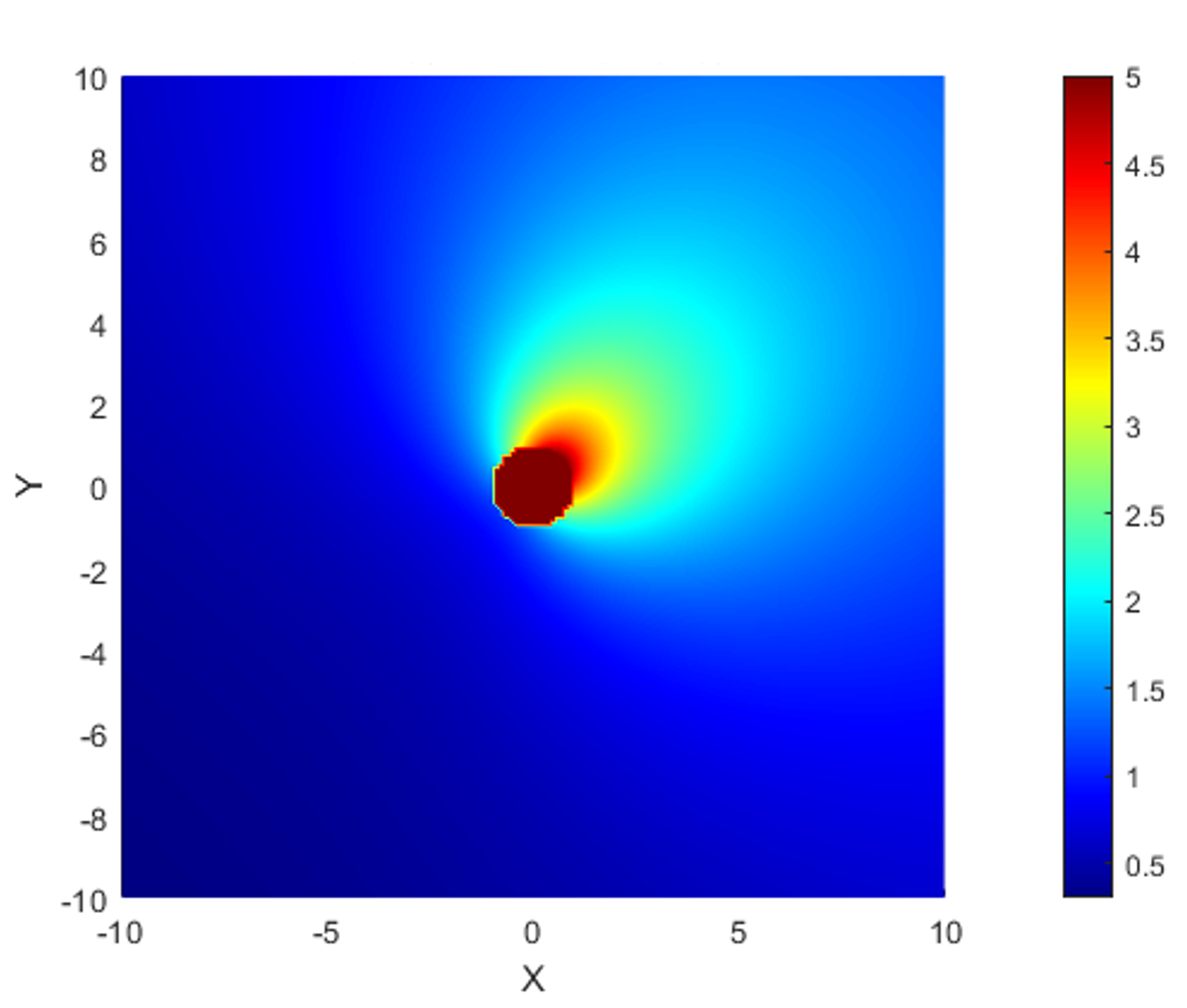}}
\caption{$(a)$ is the risk illustration of the static obstacle, its center is $(0,0)$ and its width and length are 5. $(b)$ is the risk illustration of the dynamic obstacle, its center is $(0,0)$ and its width and length are 2. The risk along the velocity of the dynamic obstacle is higher than the opposite direction.}
\label{risk_sta&dyn}
\end{figure}

To model the risk of static obstacles, we construct a spatial risk field using a two-dimensional Gaussian distribution. For the $i$-th obstacle, the risk and its gradient are defined as follows:
\begin{flalign}
R_{si}(\boldsymbol{p}_s) =\ 
&k_i \cdot \frac{1}{\sqrt{(2\pi)^3 \det(\boldsymbol{\Sigma}_i)}} \notag \\
&\cdot \exp\left( 
    -\frac{1}{2} 
    (\boldsymbol{p}_s - \boldsymbol{\mu}_i)^T 
    \boldsymbol{\Sigma}_i^{-1} 
    (\boldsymbol{p}_s -\boldsymbol{\mu}_i)
\right)&
\end{flalign}

\begin{equation}
\nabla R_{si}(\boldsymbol{p}_s) = -R_{si} \cdot \boldsymbol{\Sigma}_i^{-1} (\boldsymbol{p}_s - \boldsymbol{\mu}_i)
\end{equation}
\begin{equation}
\boldsymbol{\Sigma}_i = \begin{bmatrix}
\sigma_x^2 & 0 \\
0 & \sigma_y^2
\end{bmatrix}
\end{equation}
where, $R_{si}(\boldsymbol{p}_s)$ represents the risk field induced by static obstacles, and $\boldsymbol{\mu}_i$ denotes the center position of the $i$-th static obstacle. The matrix $\boldsymbol{\Sigma}_i$ is the covariance matrix, with its diagonal elements characterizing the obstacle's size along different directions. $\det(\boldsymbol{\Sigma}_i)$ denotes the determinant, and $\boldsymbol{\Sigma}_i^{-1}$ is the inverse of the covariance matrix. The variable $\boldsymbol{p}_s$ refers to a point in the environment. $k_i$ is the semantic weight to modulate the risk field based on obstacle semantics. A structural map alone is insufficient to differentiate between Type-II and Type-III obstacles. Through semantic analysis provided by LMSCNet, static obstacles in the environment can be classified into two categories: permanently static (e.g., trees, walls) and temporarily static (e.g., parked vehicles, standing pedestrians). To account for their higher potential risk in dynamic environments, temporarily static obstacles are assigned increased semantic weight.

\subsection{Risk Model for Dynamic Obstacles}
When a Type-III obstacle begins to move, it behaves similarly to a Type-II dynamic obstacle. In such cases, the associated risk field depends not only on the distance to the obstacle but also on its velocity. Generally, a faster-moving dynamic obstacle poses a greater risk, especially when it is closer to the agent.

To capture these characteristics, we design a dynamic obstacle risk model that accounts for both obstacle position and obstacle velocity. The model is defined as follows:
\begin{equation}R_{Di}(\boldsymbol{p}_{s})=k_d  \cdot \frac{1}{\left\|\boldsymbol{r}\right\|^{2}} \cdot \exp(k_{1}\frac{\boldsymbol{v}\cdot \boldsymbol{r}}{\left\|r\right\|})\end{equation}
\begin{equation}\nabla R_{D_i}(\boldsymbol{p}_{s})=-\frac{R_{D_i}}{\|\boldsymbol{r}\|}\cdot\left[2\cdot \boldsymbol{r}-k_{1}\cdot \boldsymbol{v}+k_{1}(\boldsymbol{v}\cdot\boldsymbol{r})\cdot\boldsymbol{r}\right]\end{equation}
\begin{equation}\boldsymbol{r}=\boldsymbol{p}_{s}-\boldsymbol{\mu}\end{equation}
where, $R_{D_i}$ denotes the risk field induced by dynamic obstacles. $\boldsymbol{r}$ is the position vector pointing from the obstacle to the aerial robot, while $k_d$ is a semantic weight that accounts for the influence of obstacle category. $\boldsymbol{v}$ denotes the velocity of the dynamic obstacle. The coefficients $k_1$ is the weight associated with velocity.

\section{Risk-aware Planning for Aerial Robot}
\label{planning}
This section presents the aerial robot planning framework, which consists of a front-end risk-informed path search and a back-end gradient-based trajectory optimization. In the front-end stage, the above mentioned risk field is incorporated to guide the path away from potentially high-risk obstacles, balancing path length with safety. The risk field ensures that the aerial robot maintains sufficient in the event of obstacle motion, thereby improving robustness in dynamic environments. In the back-end stage, we perform gradient-based trajectory optimization that accounts for multiple constraints, including dynamic feasibility, collision penalty, and trajectory smoothness, to refine the initial path into a dynamically feasible and safe trajectory.

\subsection{Path Searching Based on Risk Field and Gradient}
Traditional A* algorithms primarily focus on minimizing path length, often neglecting the potential risks in complex environments. In scenarios involving mobile obstacles, this shortest-path strategy may lead the aerial robot to fly dangerously close to obstacle surfaces. When such obstacles begin to move, the limited reaction time due to the close distance can result in collisions.

A common solution is to uniformly inflate all obstacle boundaries. However, this approach ignores the semantic differences among obstacles and treats low-risk static obstacles the same as potentially dynamic ones. Consequently, the aerial robot unnecessarily avoids all obstacles, leading to excessive path deviation and loss in efficiency.

To address this issue, we propose a risk-informed A* path planning algorithm (R-A*) that considers both risk value and corresponding gradient, leveraging semantic information from the environment. This method not only preserves path efficiency but also enables the aerial robot to proactively avoid high-risk areas. When navigating near risk areas, the path is intelligently guided by the gradient direction to shift toward lower-risk areas.

Assume the environment contains $M$ static and $N$ dynamic obstacles. The overall risk field is constructed by superimposing the individual risk from all obstacles. The resulting unified risk field $R({\boldsymbol{p}_s})$ and its gradient $\nabla R({\boldsymbol{p}_s})$ are then used to guide the path search process.
\begin{equation}
    R(\boldsymbol{p}_{s}) = \sum_{i = 1}^{M} R_{Si}(\boldsymbol{p}_{s}) + \sum_{i = 1}^{N} R_{Di}(\boldsymbol{p}_{s})
\end{equation}
\begin{equation}
    \nabla R(\boldsymbol{p}_s) = \sum_{i = 1}^{M} \nabla R_{Si}(\boldsymbol{p}_s) + \sum_{i = 1}^{N} \nabla R_{Di}(\boldsymbol{p}_s)
\end{equation}

The gradient of the risk field indicates the direction in which the risk increases most rapidly. Therefore, in path planning, the search direction should be adjusted along the negative gradient to avoid high-risk areas. To incorporate risk into the planning process, we design a cost function that jointly considers both the risk and gradient, thereby enabling risk-sensitive navigation behavior.

Unlike the traditional A* algorithm, which evaluates nodes based on heuristic distance, we define the cost of each candidate node 
$\boldsymbol{p}_{(id)}$ as:
\begin{equation}
    f(\boldsymbol{p}_{id}) = g(\boldsymbol{p}_{id}) + h(\boldsymbol{p}_{id}) + \lambda R(\boldsymbol{p}_{id}) + \alpha G(\boldsymbol{p}_{id})
\end{equation}
where, $g(\boldsymbol{p}_{id})$ denotes the accumulated cost from the start node to the current node, and $h(\boldsymbol{p}_{id})$ is the heuristic estimate from the current node to the goal, computed as the Euclidean distance. $\lambda$ is the weighting coefficient for the risk value, and $R(\boldsymbol{p}_{id})$ is the risk value obtained from the Gaussian-based risk field. $\alpha$ is the weight for the gradient guidance term, and $G(\boldsymbol{p}_{id})$ represents the risk gradient component, which guides the aerial robot away from high-risk areas. By integrating both the risk value and gradient into the cost function, the path search algorithm can intelligently adjust the planning direction when approaching obstacles, thereby avoiding high-risk obstacles and improving path safety and robustness.

However, when a previously static obstacle begins to move, the corresponding risk field becomes dynamically asymmetric. As shown in Fig.\ref{fig:b1}, the side in the direction of the obstacle's velocity exhibits a significantly higher risk value than the opposite side. In such cases, relying solely on the gradient along one side of the risk field can cause the planner to become trapped in local minima, causing it to fail in avoiding the dynamic obstacle effectively.
\begin{figure}
    \centering
    \includegraphics[width=1.0\linewidth]{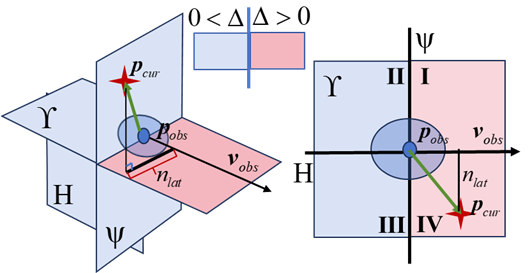}
    \caption{Local coordinate system centered at the obstacle. When $\Delta > 0$, the position lies in front (quadrants I and IV); when $\Delta < 0$, the position is behind (quadrants II and III).
}
    \label{delta}
\end{figure}

To address this limitation, we establish a local coordinate system centered at the obstacle's current position $\boldsymbol{p}_{obj}$, with the velocity vector $\boldsymbol{v}_{obj}$ defined as the positive direction of the primary axis, as illustrated in the Fig.\ref{delta}. Based on this coordinate frame, we construct three mutually orthogonal planes to characterize the spatial influence of the moving obstacle. Specifically, $\mathbf{\Psi}$ denotes the plane perpendicular to the velocity vector $\boldsymbol{v}_{obj}$ and passing through the obstacle center; $\mathbf{\Upsilon}$ represents the horizontal plane on which the obstacle resides; and $\mathbf{H}$ is the vertical plane perpendicular to $\mathbf{\Upsilon}$ and containing $\boldsymbol{v}_{obj}$. Building upon the local coordinate system, we introduce a direction factor $\Delta$, which enables more informed directional adjustments during motion planning. This directional factor improves the planner’s high-risk obstacle avoidance capability by biasing the path toward directions that account for both the obstacle’s velocity and its position. The formulation of $\Delta$ is given as
\begin{equation}
    \Delta = \frac{(\boldsymbol{p}_{curr} - \boldsymbol{p}_{obs}) \cdot \boldsymbol{v}_{obs}}{\|\boldsymbol{v}_{obs}\| + \varepsilon}
\end{equation}

When $\Delta > 0$, the current position lies in the forward direction of the obstacle motion, corresponding to the first or fourth quadrant in the local coordinate system. Conversely, when $\Delta < 0$, the position lies behind the obstacle, within the second or third quadrant.
Based on the quadrant in which the current node is located, we adjust the risk gradient guidance term $G_({\boldsymbol{p}_{id}})$ to navigate toward safer and more reasonable directions for obstacle avoidance.
\begin{equation}
G(\boldsymbol{p}_{id}) = 
\begin{cases}
\nabla R(\boldsymbol{p}_{id}) \cdot \boldsymbol{d}_{air}, & \text{if } \Delta < 0 \\
\boldsymbol{v}_{obs} \cdot \boldsymbol{d}_{air}, & \text{if } \Delta > 0 \text{ and } n_{last} \geq n_{ref} \\
\nabla R(\boldsymbol{p}_{id}) \cdot \boldsymbol{d}_{air}, & \text{if } \Delta > 0 \text{ and } n_{last} < n_{ref}
\end{cases}
\end{equation}

\begin{equation}
    \boldsymbol{d}_{air} = \boldsymbol{p}_{id} - \boldsymbol{p}_{curr}
\end{equation}
Here, $\boldsymbol{p}_{id}$ denotes the next neighboring node, $\boldsymbol{p}_{curr}$ denotes the current node in the path search process. $ n_{last}$ represents the projection of the distance from $\boldsymbol{p}_{cur}$ to $\boldsymbol{v}_{obs}$ onto the plane $\mathbf{\Upsilon}$. $ n_{ref}$ is a predefined safety reference distance.
When the position lies behind the moving obstacle, the planner follows the descending direction of the risk gradient, which naturally directs the path away from high-risk areas. If the position is in front of the obstacle but at a relatively long distance $( n_{last} \geq n_{ref})$, the planner prefers to bypass the obstacle from the rear. In contrast, when the position is in front and close to the obstacle $( n_{last} < n_{ref})$, a forward avoidance path tends to have a lower cost and is therefore more likely to be chosen. The directional factor $\Delta$ prevents the planner from being confined to one-sided gradient descent, avoiding local minima and enhancing its global effectiveness in dynamic environments.

\begin{algorithm}[t]
\caption{Risk-Informed A* Path Search with Directional Factor}
\begin{algorithmic}[1]
\State \textbf{Input:} Grid map $\mathcal{M}$, semantic-based risk field $R(\boldsymbol{p}_{id})$, risk gradient field $\nabla R(\boldsymbol{p}_{id})$
\State \textbf{Output:} A safe path $\mathcal{P}$
\State Initialize open list $\mathcal{O} \gets \{\boldsymbol{p}_{\text{start}}\}$, closed list $\mathcal{C} \gets \emptyset$
\State $g(\boldsymbol{p}_{id}) \gets \infty, \ \forall n$, $g(\boldsymbol{p}_{\text{start}}) \gets 0$
\State $parent(\boldsymbol{p}_{id}) \gets$ None, $\forall \boldsymbol{p}_{id}$
\While{$\mathcal{O} \neq \emptyset$}
    \State Select $\boldsymbol{p}_{\text{curr}} \in \mathcal{O}$ with minimal $f(\boldsymbol{p}_{id})$
    \If{$\boldsymbol{p}_{\text{curr}} = \boldsymbol{p}_{\text{goal}}$}
        \State \Return RetrievePath($\boldsymbol{p}_{\text{goal}}$)
    \EndIf
    \State Move $\boldsymbol{p}_{\text{curr}}$ from $\mathcal{O}$ to $\mathcal{C}$
    \ForAll{neighbor $\boldsymbol{p}_{\text{next}}$ of $\boldsymbol{p}_{\text{curr}}$}
        \If{$\boldsymbol{p}_{\text{next}} \in \mathcal{C}$}
            \State \textbf{continue}
        \EndIf
        \State $g_{\text{temp}} \gets g(\boldsymbol{p}_{\text{curr}}) + \text{MoveCost}(\boldsymbol{p}_{\text{curr}}, \boldsymbol{p}_{\text{next}})$
        \If{$g_{\text{temp}} < g(\boldsymbol{p}_{\text{next}})$}
            \State $g(\boldsymbol{p}_{\text{next}}) \gets g_{\text{temp}}$
            \State $parent(\boldsymbol{p}_{\text{next}}) \gets \boldsymbol{p}_{\text{curr}}$
        \EndIf
        \State $h(\boldsymbol{p}_{\text{next}}) \gets \| \boldsymbol{p}_{{\text{next}}} - \boldsymbol{p}_{\text{goal}} \|$
        \State Get $R(\boldsymbol{p}_{\text{next}})$ from risk field
        \If{$\boldsymbol{p}_{\text{next}}$ is near a dynamic obstacle}
            \State $\Delta \gets \frac{(\boldsymbol{p}_{curr} - \boldsymbol{p}_{obs}) \cdot \boldsymbol{v}_{obs}}{\|\boldsymbol{v}_{obs}\| + \varepsilon}$
            \State Determine front/rear region based on $\Delta$
            \State Adjust $G(\boldsymbol{p}_{\text{next}})$ according equation (11)
        \Else
            \State $G(\boldsymbol{p}_{\text{next}}) \gets -\nabla R(\boldsymbol{p}_{\text{next}})$
        \EndIf
        \State $f(\boldsymbol{p}_{\text{next}}) \gets g(\boldsymbol{p}_{\text{next}}) + h(\boldsymbol{p}_{\text{next}}) + \lambda R(\boldsymbol{p}_{\text{next}}) + \alpha G(v_{\text{next}})$
        \If{$\boldsymbol{p}_{\text{next}} \notin \mathcal{O}$}
            \State Add $\boldsymbol{p}_{\text{next}}$ to $\mathcal{O}$
        \EndIf
    \EndFor
\EndWhile
\State \Return Failure (no feasible path found)
\end{algorithmic}
\end{algorithm}

\subsection{Trajectory Optimization}
In this paper, the initial trajectory of the aerial robot is represented by a B-spline. The B-spline has an order of $p_b$, defined by $N_c$ control points $\{\boldsymbol{Q}_1,\boldsymbol{Q}_2,\cdots,\boldsymbol{Q}_{N_c}\}$ and a knot vector  $\{t_1,t_2,\cdots,t_{M_b}\}$, where $\boldsymbol{Q}_{i} \in R^3, t_m \in R $ and $M_b=N_{c}+p_{b}$. For the efficiency of trajectory evaluation, we use $p_b$-order uniform B-spline $\Phi \left( t\right)$ to represent the initial trajectory of the drone. For a uniform B-spline, every node interval has the same value $\Delta t=t_{m+1}-t_{m}$.Since $\Delta t$ is identical along $\Phi \left( t\right)$, and the derivative of the B-spline is still a B-spline curve, and then the velocity $\boldsymbol{V}_{i}$ and acceleration $\boldsymbol{A}_{i}$ and jerk $\boldsymbol{J}_{i}$ of the control point are obtained by

\begin{equation}
\boldsymbol{V}_{i} = \frac{\boldsymbol{Q}_{i + 1} - \boldsymbol{Q}_{i}}{\Delta t},\boldsymbol{A}_{i} = \frac{\boldsymbol{V}_{i + 1} - \boldsymbol{V}_{i}}{\Delta t},\boldsymbol{J}_{i} = \frac{\boldsymbol{A}_{i + 1} - \boldsymbol{A}_{i}}{\Delta t}.
\end{equation}

\begin{figure}
    \centering
    \includegraphics[width=1.0\linewidth]{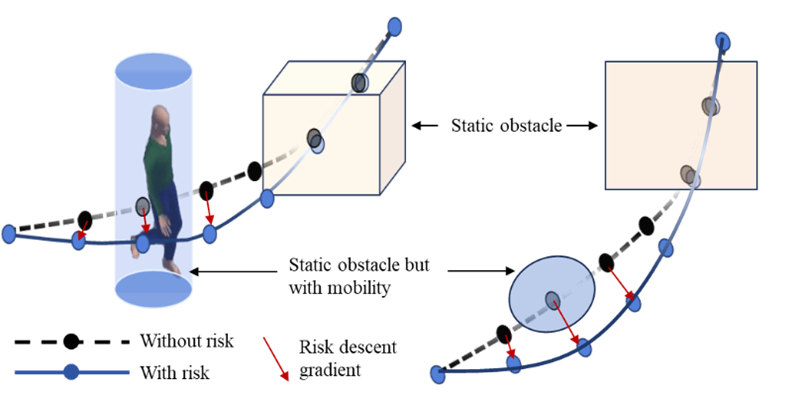}
    \caption{Initialization of control points. When the risk value of an obstacle exceeds the threshold $R_{thresh}$, the control points are shifted along the direction of risk reduction.}
    \label{controlPoint}
\end{figure}

We utilize the aforementioned semantic segmentation to identify such static obstacles with potential mobility and to obtain their risk gradients. During the initialization of the control points $\boldsymbol{Q}_{i}$, a displacement $r_d$ is applied in the direction of the risk gradient descent, as shown in Fig.\ref{controlPoint}.The control points are initialized as follows:
\begin{equation}
\boldsymbol{Q}_i =
\begin{cases}
\boldsymbol{Q}_i - \dfrac{\nabla R(\boldsymbol{Q}_i)}{\|\nabla R(\boldsymbol{Q}_i)\|} \cdot r_d, & \text{if } R(\boldsymbol{Q}_i) \ge R_{\text{thresh}}, \\[8pt]
\boldsymbol{Q}_i, & \text{if } R(\boldsymbol{Q}_i) < R_{thresh}.
\end{cases}
\end{equation}

We treat the control points $\boldsymbol{Q}$ of the uniform B-spline $\Phi(t)$ as the optimization variables. The aerial robot trajectory is optimized by simultaneously considering trajectory smoothness, collision avoidance, and dynamic feasibility, leading to a gradient-based unconstrained objective function:

\begin{equation}
\min_{\boldsymbol{Q}} J = \lambda_s J_s + \lambda_c J_c + \lambda_d J_d + \lambda_r J_r,
\end{equation}
where $\lambda_s, \lambda_c, \lambda_d,$ and $\lambda_r$ are the respective weights for each penalty term. Here, $J_s$ represents the smoothness cost, $J_c$ accounts for static collision avoidance, $J_d$ enforces dynamic feasibility, and $J_r$ penalizes dynamic collision risks. 
The smoothness of the trajectory is ensured by leveraging the control points of the higher-order derivatives of the B-spline, exploiting the convex hull property.
\begin{equation}
{{J}}_{s} = \mathop{\sum }\limits_{{i = 1}}^{{{N}_{c} - 1}}{\begin{Vmatrix}{\boldsymbol{A}}_{i}\end{Vmatrix}}_{2}^{2} + \mathop{\sum }\limits_{{i = 1}}^{{{N}_{c} - 2}}{\begin{Vmatrix}{\boldsymbol{J}}_{i}\end{Vmatrix}}_{2}^{2}.
\end{equation}

The collision penalty enforces avoidance of static obstacles by generating repulsive forces that push the aerial robot away. When the initial aerial robot trajectory passes through an obstacle, global safe guide path is obtained by R-A* to avoid collisions. 

\begin{equation}
    J_c = \sum_{i=1}^{N_c} \sum_{j=1}^{N_i} \max \left\{d_{ij}-s_f ,0\right\}^3,
\end{equation}
where, $s_f$ is safety clearance, $d_{ij}$ is the obstacle distance between $\boldsymbol{Q}_i$ and the $j$-th obstacle, see details in \cite{c4}. $N_i$ is the number of $d_{ij}$ belonging to $\boldsymbol{Q}_i$.

Dynamic feasibility is ensured by constraining the aerial robot's velocity and acceleration within predefined limits.
\begin{equation}  
J_d = \sum_{j=1}^{N_c-1} g\Big(\|\boldsymbol{V}_j\|^2 - v_m^2\Big) 
    + \sum_{j=1}^{N_c-2} g\Big(\|\boldsymbol{A}_j\|^2 - a_m^2\Big),
\end{equation}
where, $v_m$ and $a_m$ are the maximum velocity and acceleration,
$g(\cdot)$ is the $\mathbf{C}^2$ penalty function $g(x) = \max(0,x)^3$.

For dynamic high-risk obstacles, we define the following penalty function $J_r$:
\begin{equation}
J_r =
\begin{cases}
0, & R(\boldsymbol{Q}_i) < R_{\text{thresh}} \\
\displaystyle \sum_{i=1}^{N_c} \sum_{j=1}^{N_d} g \Big(C - \big(\boldsymbol{Q}_i - \boldsymbol{P}_j \big)\Big), & R(\boldsymbol{Q}_i) \geq R_{\text{thresh}}
\end{cases}
\end{equation}

\begin{equation}
   \boldsymbol{P}_j= \boldsymbol{p}_{{obs},j} - \boldsymbol{v}_{{obs},j}\cdot \Delta t \cdot i
\end{equation}
where $N_d$ denotes the number of dynamic obstacles, and $C$ represents the safety distance threshold. 
The term $\boldsymbol{P}_j$ corresponds to the predicted future trajectory of the $j$-th dynamic obstacle, 
where $p_{{obs},j}$ is the current obstacle position obtained through point cloud clustering, 
and $v_{{obs},j}$ is the obstacle velocity estimated by point cloud registration across frames.

\section{Experiment and Results}
\label{simulation}
In this section, we first evaluate the performance of the semantic segmentation network on the SemanticKITTI dataset. Based on semantic perception, we then evaluate the R-A* path’s performance, with particular attention to its distance from high-risk obstacles and the length of the planned paths. Furthermore, simulation experiments are conducted to assess the capability of the risk-aware planner in unknown environments, with particular attention to planning success rate, average flight time, and average planning time. Finally, simulation tests based on real-world environmental data are carried out to validate the proposed system.

\subsection{Experiment Setup}
For the semantic perception module, we trained and tested LMSCNet on the SemanticKITTI dataset \cite{c28} using an NVIDIA RTX 4090 GPU. The training was conducted for 80 epochs with a batch size of 4, employing the Adam optimizer with an initial learning rate of 0.0005.

In the simulation experiments, two types of evaluations were carried out. Path search tests were conducted in MATLAB on randomly generated 50× 50 2D grid maps in a dimensionless setup, where dynamic high-risk obstacles with motion capability were predefined (as shown in Fig.\ref{path search}). In this scenario, the aerial robot is required to fly from the start point to the goal while avoiding collisions. Dynamic planning tests were performed in a Gazebo + ROS1 environment, where a realistic urban scene is built, including persons, vehicles, trees, and buildings. Among the four pedestrians, two had the ability to switch from static state to moving state: one was visible within the aerial robot’s field of view, while the other was occluded by a building. These persons move along the aerial robot’s flight direction, and the aerial robot need to navigate safely from start to goal while avoiding dynamic risks.

\subsection{LMSCNet Segmentation Performance}
We evaluate LMSCNet based on the state-of-the-art SSC methods on the SemanticKITTI test dataset. As shown in Table~\ref{tab:segmentation_comparison}, LMSCNet not only achieves a high completion metric IoU (58.76\%) but also ranks first in the semantic segmentation metric mIoU (17.01\%). While SCPNet attains higher semantic segmentation accuracy than our approach, its densely structured network limits real-time inference. By comparison, LMSCNet not only surpasses SCPNet by 4.74$\%$ in IoU but also achieves an inference speed nearly 13 times faster.

\begin{table}[htbp]
\centering
\caption{Comparison of different methods on semantic segmentation metrics}
\resizebox{\linewidth}{!}{\begin{tabular}{lcccc}
\hline
\textbf{Method} & \textbf{IoU} & \textbf{mIoU} & \textbf{Prec.} & \textbf{FPS} \\
\hline
SCPNet\cite{c31}        & 56.10 & 36.70 & 72.43 & $<$1 \\
VoxFromer-S\cite{c30}   & 57.54 & 16.48 & 70.85 & $<$1 \\
SSCNet\cite{c32}        & 53.20 & 14.55 & 59.13 & 12.00 \\
LMSCNet\cite{c21}       & 55.32 & 17.01 & 77.11 & \textbf{13.50} \\
\hline
\end{tabular}}
\label{tab:segmentation_comparison}
\end{table}

\subsection{R-A* Path Search Performance}
\begin{figure}
    \centering
    \includegraphics[width=0.9\linewidth]{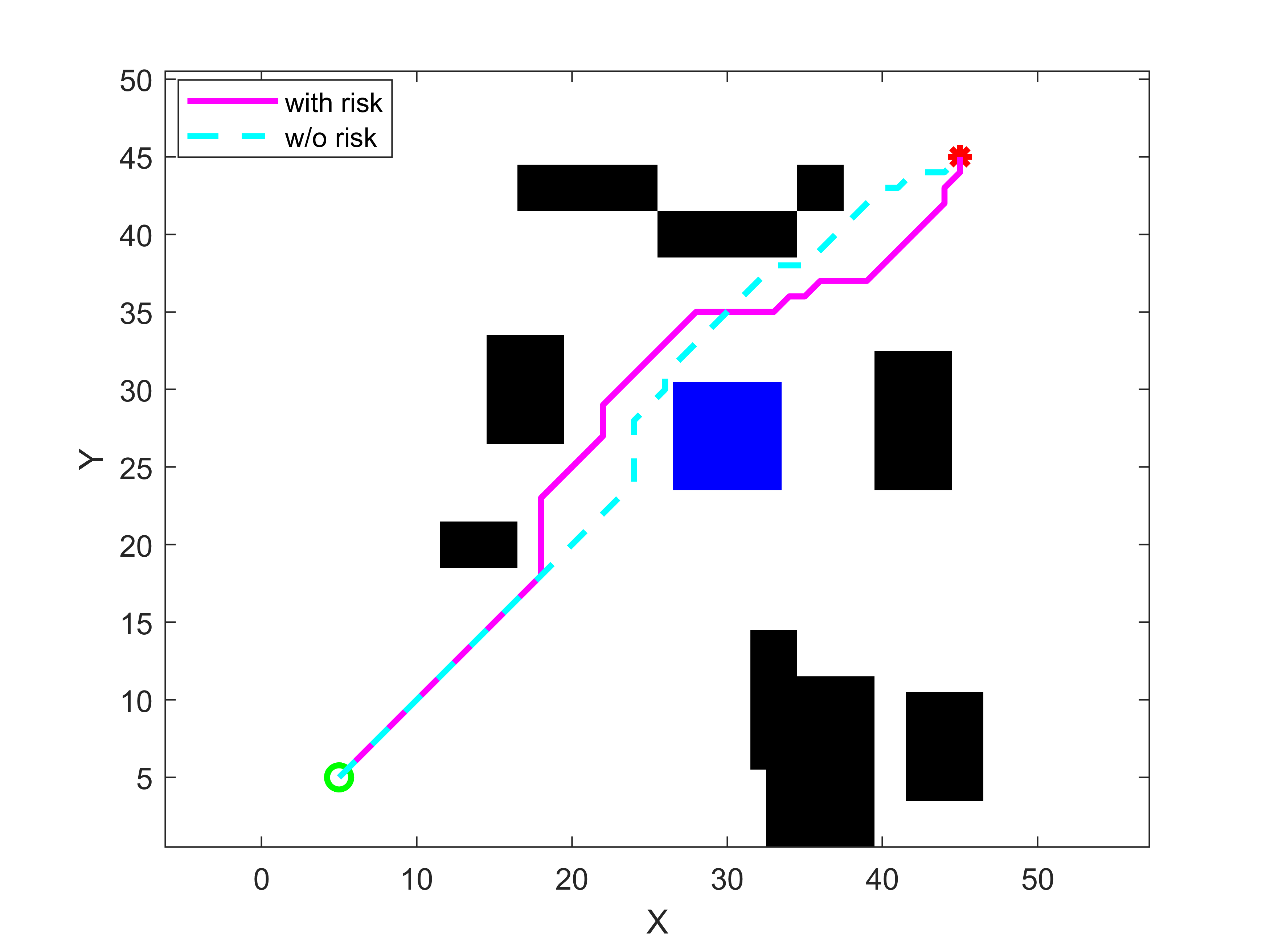}
    \caption{Comparison of A* path search results in high-risk static environment. The random dimensionless map size is 50×50, where black indicates static obstacles, blue denotes predefined high-risk obstacles with motion capability, the green point represents the start, and the red point represents the goal.}
    \label{path search}
\end{figure}

\begin{figure}[!t]
\centering
\subfloat[\label{fig:a}]{
		\includegraphics[scale=0.25]{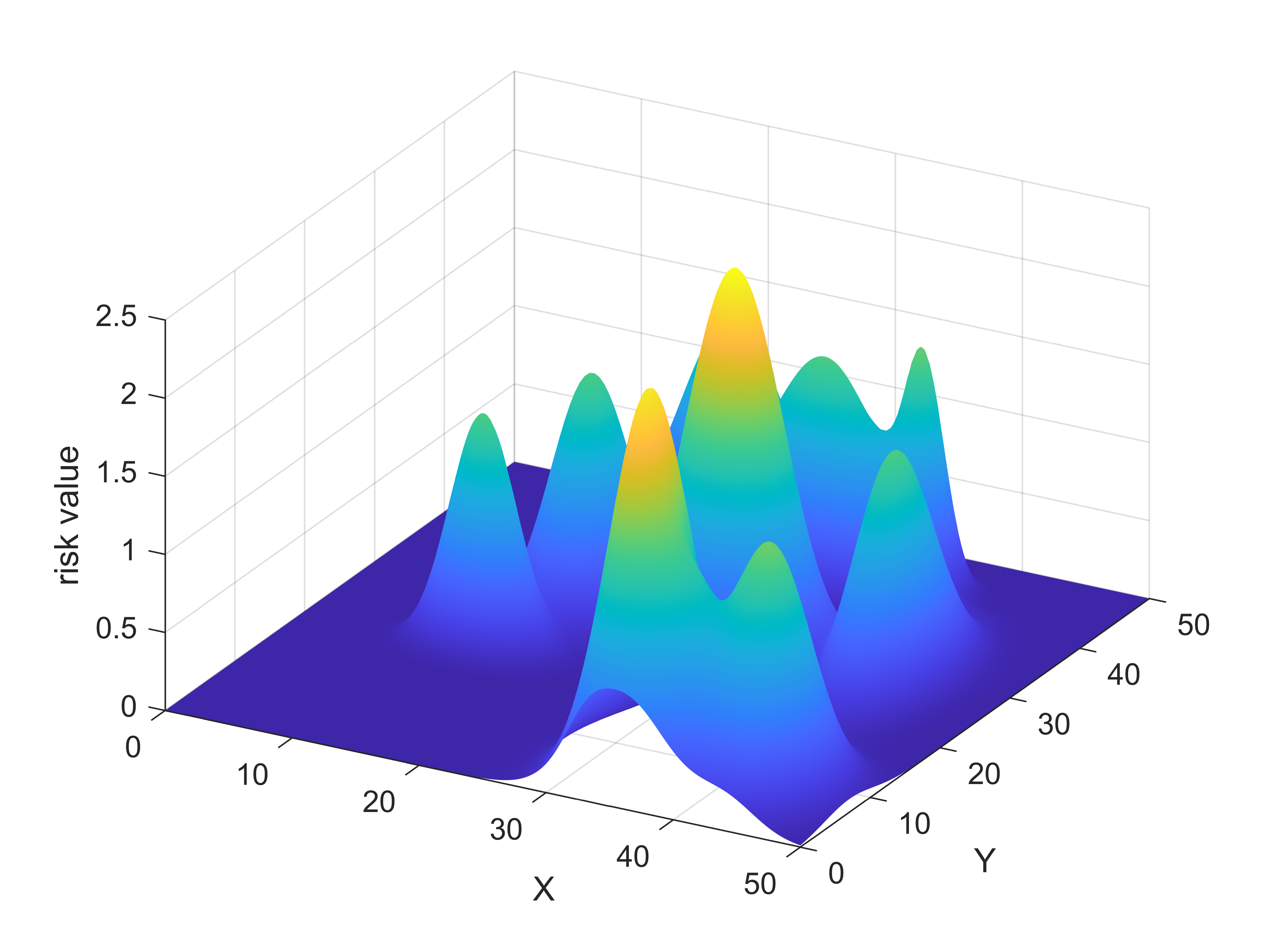}}
\subfloat[\label{fig:b}]{
		\includegraphics[scale=0.25]{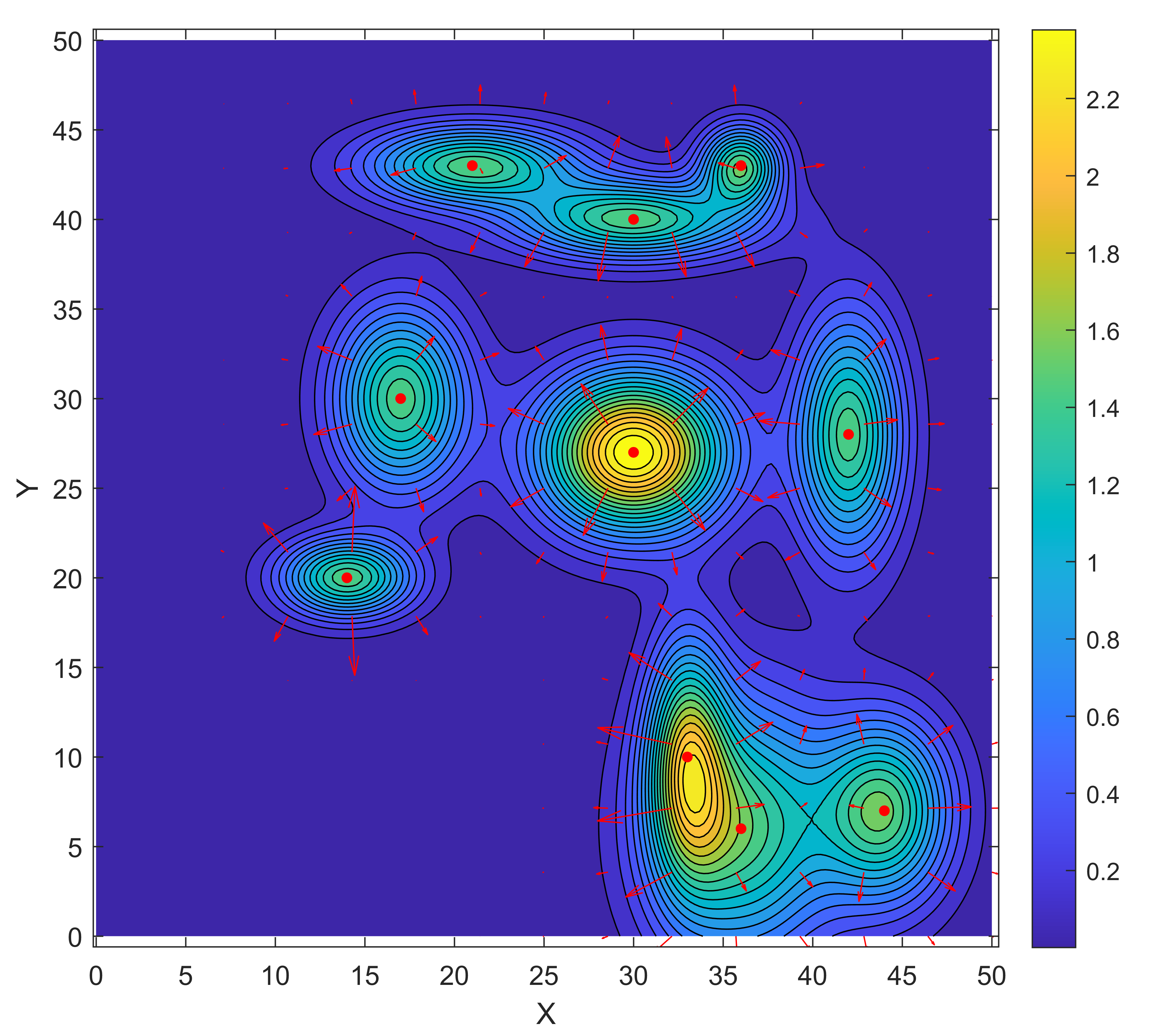}}
\caption{Risk potential field and corresponding risk gradient of the random map, where red arrows indicate the direction of risk gradient descent.}
\label{risk}
\end{figure}

In the experiment, a random 50×50 dimensionless map was generated, in which obstacles with motion capability were predefined and marked in blue (Fig.\ref{path search}). Based on the risk estimation function for high-risk static obstacles, we further visualized the superimposed risk potential field (Fig.\ref{fig:a}) and computed the corresponding risk gradient of the map (Fig.\ref{fig:b}). The results clearly show that obstacles with motion capability exhibit significantly higher risk values than static ones.

On this random map, we evaluated the proposed R-A* path search algorithm and compared it with the traditional A* algorithm. As illustrated in Fig.\ref{path search}, the red solid line denotes the risk-aware path, while the blue dashed line represents the path obtained by the traditional A* algorithm. It can be observed that the traditional A* algorithm, lacking risk awareness, tends to generate paths that are close to obstacles. In contrast, the R-A* algorithm incorporates both risk values and risk gradients during path searching, thereby balancing path length and safety. Specifically, it prefers routes that stay close to low-risk static obstacles while keeping a safe distance from high-risk dynamic obstacles.

Quantitative results show that the risk-aware path has a length of $61.8$, compared to $59.5$ for the traditional A* path, with only a $3.9\%$ increase. Meanwhile, the minimum distance between the risk-aware path and high-risk obstacles is 2.83, whereas the traditional A* path approaches as close as 1. These findings demonstrate that the proposed R-A* method effectively enhances path safety while maintaining near-optimal path length, achieving risk avoidance at a minimal additional cost.

\begin{table}[htbp]
    \centering
    \caption{Path search result comparison in high-risk environment}
   \resizebox{\linewidth}{!}{ \begin{tabular}{lccc}
     \toprule
        Scene & Method & Path Length & Minimum Safe Distance\\
     \midrule
        \multirow{2}{*}{Static Env.} 
          & R-A*(ours)   & 61.8  & 2.83 \\
          & A*\cite{c23}   & 59.5  & 1    \\
     \midrule  
        \multirow{3}{*}{Dynamic Env.} 
          & R-A*(ours)   & 65.25 & 2.06 \\
          & A*\cite{c23}   & 62.33 & 0    \\
          & D*\cite{c24}   & 66.33 & 0.5  \\      
    \bottomrule
    \end{tabular}}
    \label{path search vs}
\end{table}

\begin{figure}
    \centering
    \includegraphics[width=1.0\linewidth]{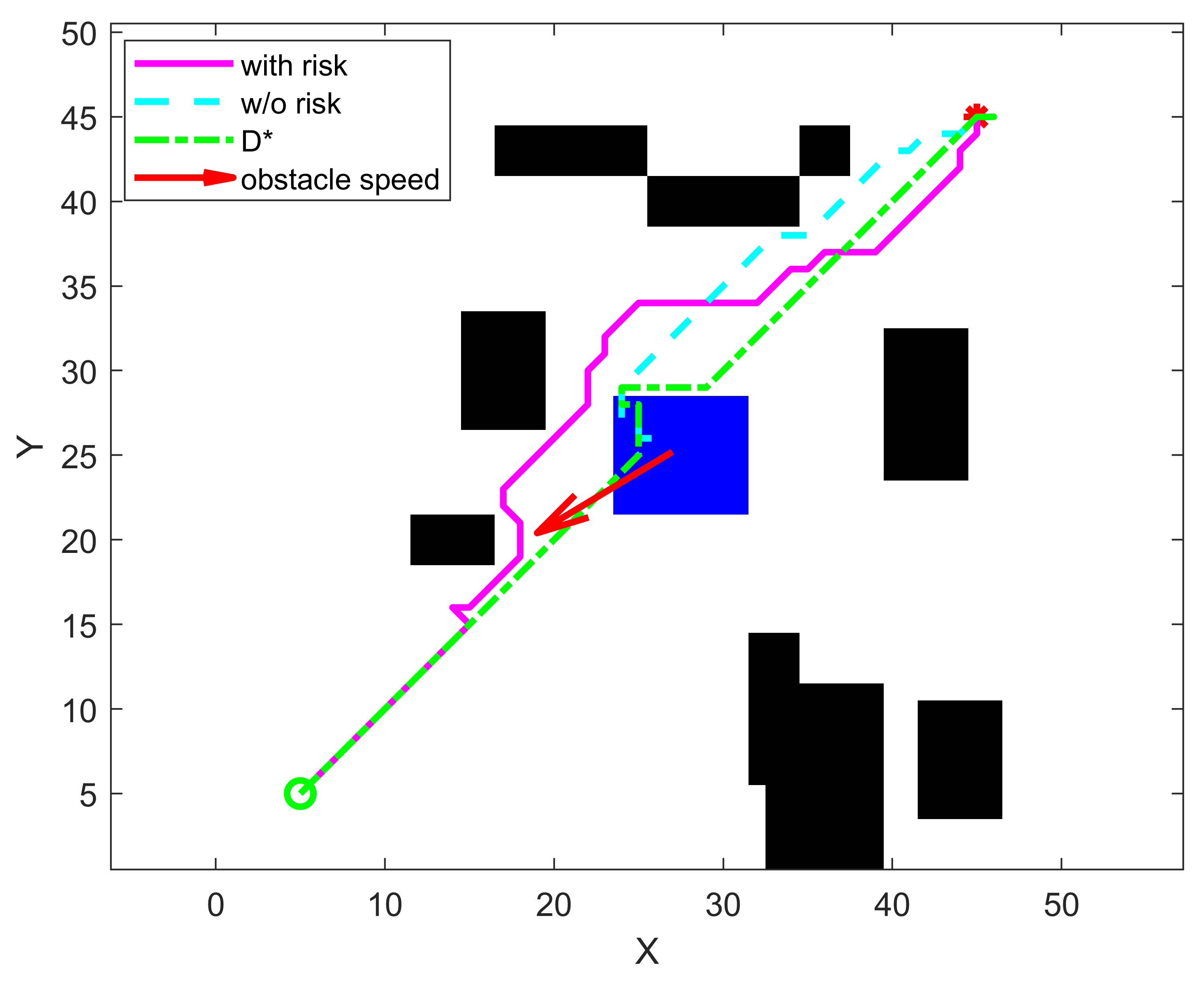}
    \caption{Comparison of path search results in high-risk dynamic environment. When a search node approaches a moving obstacle (marked in blue), the obstacle moves with a velocity of $(-0.5, -0.3)$.}
    \label{dynamicvs}
\end{figure}

To assess the responsiveness of the path planning algorithm in dynamic obstacle scenarios, when a search node approaches an obstacle within the safety threshold $d_\text{safe}$, the obstacle moves with a velocity of $(-0.5, -0.3)$, altering the environmental risk field (see Fig.\ref{dynamicvs}). Under these conditions, the proposed R-A* is compared with conventional A* and D* algorithms.

Results demonstrate that R-A* proactively avoids high-risk obstacles, effectively reducing collision risk. As summarized in Table\ref{path search vs}, R-A* produces a path length of 65.25, slightly longer than conventional A* (62.33) but shorter than D* (66.33). Importantly, the minimum distance between the Guided A* path and high-risk obstacles is 2.06, substantially larger than those of A* and D*, indicating that the proposed method achieves a favorable balance between safety and path efficiency.

\begin{figure}[!t]
    \centering
    \includegraphics[width=1.0\linewidth]{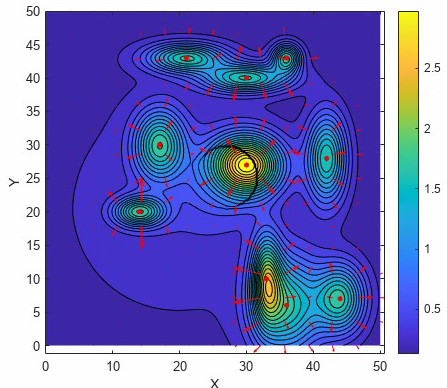}
    \caption{Environmental risk field and its gradient corresponding to the obstacle in motion.}
    \label{dyriskgrad}
\end{figure}

\subsection{Simulated High-risk Environment Planning}

\begin{figure}
    \centering
    \includegraphics[width=1.0\linewidth]{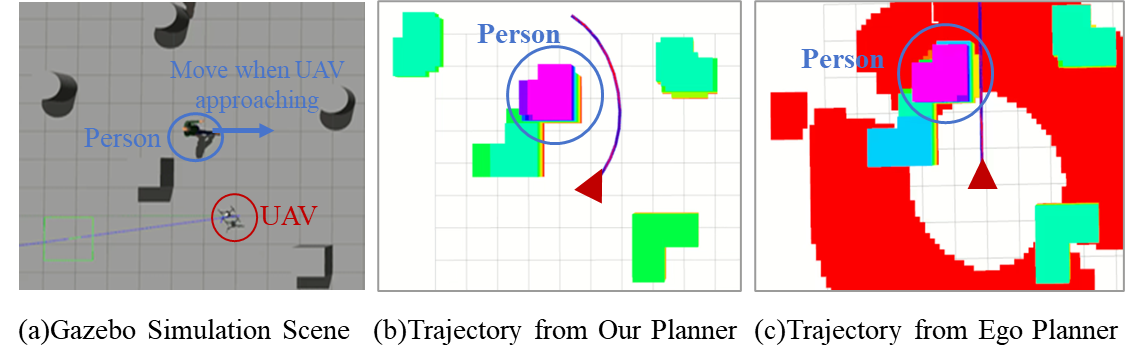}
    \caption{Comparison of trajectories with the baseline in the same scenario. (a) Simulation environment in Gazebo. (b) Our method generates a safe trajectory by accounting for potential obstacle risks. (c) Ego-Planner fails to assess such risks and produces a trajectory close to the obstacle with potential collision risk.}
    \label{scene1_static}
\end{figure}

\begin{figure}
    \centering
    \includegraphics[width=1.0\linewidth]{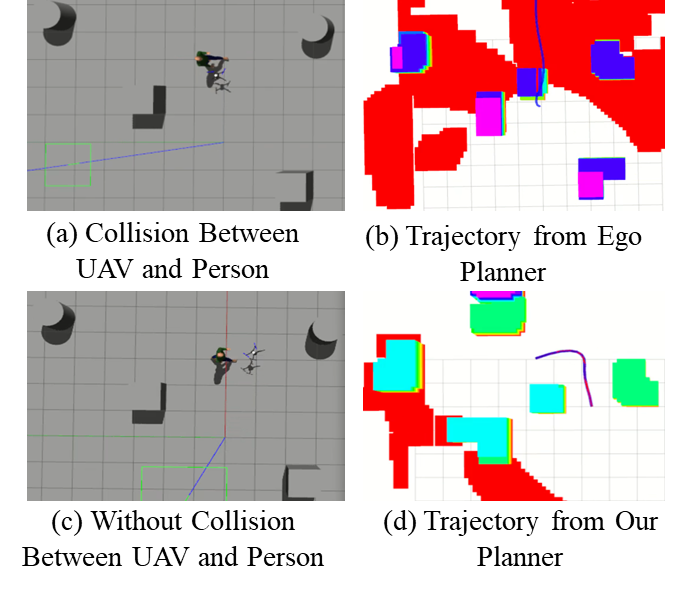}
    \caption{Trajectory comparison with the baseline in the dynamic scenario. (a) In the Gazebo simulation, the ‘person’ starts moving when the Aerial Robot approaches, resulting in a collision. (b) Ego-Planner generates an initial trajectory close to the obstacle, and when the 'person' moves, it fails to react in time, causing a collision. (c) In the Gazebo simulation, the Aerial Robot successfully avoids the 'person' without collision. (d) Our planner accounts for the potential risk of the 'person' and generates an initial trajectory away from the obstacle, enabling timely avoidance when the 'person' moves.}
    \label{scene1_dynamic}
\end{figure}

In the two proposed simulation environments, we conducted multiple comparative experiments to qualitatively analyze the advantages of our method over the baseline. Fig.\ref{scene1_static}a illustrates the first scenario where an aerial robot navigates toward the target while a 'person' crosses its path as the aerial robot approaches. Ego-Planner models obstacles as occupied grid maps, without distinguishing obstacle categories or evaluating potential risks. As a result, it generates high-risk trajectories close to obstacles(Fig.\ref{scene1_static}c). When the 'person' moves, the aerial robot lacks sufficient reaction distance and eventually collides(Fig.\ref{scene1_dynamic}b). In this simulation, we get labels for different obstacles. Based on this information, our planner considers the potential risks of different obstacle types and generates safe bypass trajectories by jointly optimizing risk avoidance and path length (Fig.\ref{scene1_static}b). Because the trajectory remains distant from obstacles, the aerial robot can avoid collisions when the 'person' moves(Fig.\ref{scene1_dynamic}d).  

\begin{figure}
    \centering
    \includegraphics[width=1.0\linewidth]{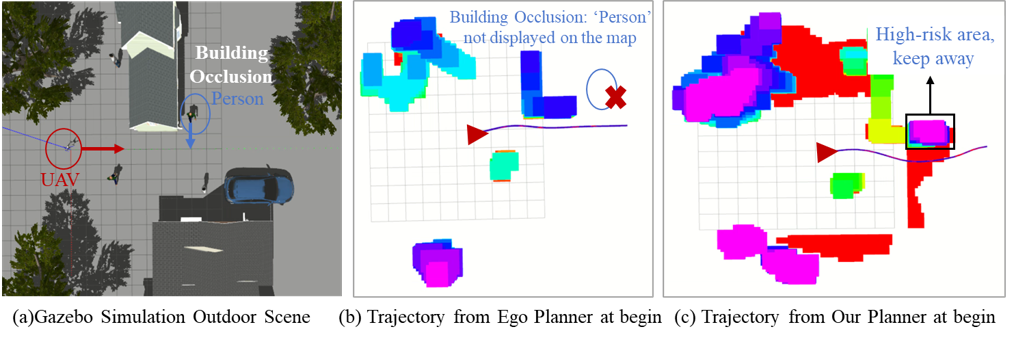}
    \caption{Trajectory comparison in the same outdoor simulation scenario. (a) In the Gazebo environment, the ‘person’ is occluded by a house and starts moving when the Aerial Robot approaches. (b) Ego-Planner lacks semantic understanding of the environment and fails to account for the occlusion risk, resulting in a trajectory close to the obstacle. (c) Our planner leverages house semantics to construct a risk field in advance and generates a safe trajectory away from the obstacle.}
    \label{scene2_static}
\end{figure}

\begin{figure}
    \centering
    \includegraphics[width=1.0\linewidth]{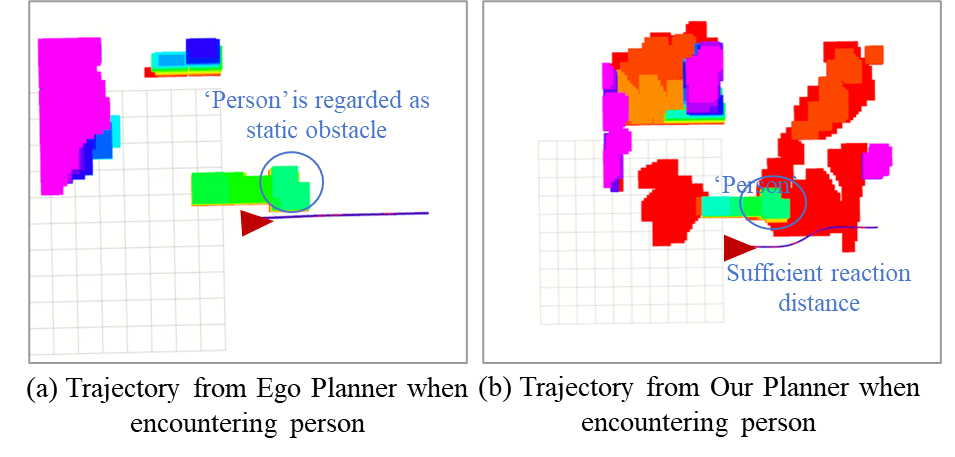}
    \caption{Trajectory comparison when the Aerial Robot encounters the ‘person’. (a) Ego-Planner generates a trajectory close to the house, leading to a collision when the ‘person’ continues to move. (b) Our planner generates a trajectory away from the house, providing sufficient reaction distance for the Aerial Robot to avoid the obstacle when the ‘person' continues to move.}
    \label{scene2_dy}
\end{figure}

Furthermore, we performed comparative tests in an outdoor simulation environment. As shown in Fig.\ref{scene2_static}a, the scenario includes static trees, humans, and houses, as well as moving humans, including one occluded by a house. When the Aerial Robot flies forward from the starting point, it encounters a person going from behind the house. As shown in Fig.\ref{scene2_static}b, Ego-Planner does not account for the risks of occlusion and thus generates trajectories close to obstacles, lacking a semantic understanding of the environment. When the 'person' goes into the detectable region, Ego-Planner models it as a static obstacle through map inflation, resulting in a trajectory that remains close to the person(Fig.\ref{scene2_dy}a). Consequently, collisions occur if the person continues to move. In contrast, our planner considers semantic information and constructs a Gaussian risk field, treating house corners as high-risk regions. This enables the generation of trajectories that steer away from the house, as shown in Fig.\ref{scene2_static}c. When the 'person' is detected, the Aerial Robot maintains sufficient safety distance to perform timely obstacle avoidance.(Fig.\ref{scene2_dy}b)

In two simulated scenes, we conducted 40 experiments each and evaluated the success rate, planning time, and flight time of different methods, as summarized in Table \ref{tab:simulation_results}. We further analyzed the contribution of different modules by testing two variants: Ours (only R-A*), which uses risk-informed A* only during path search, and Ours (full), which combines R-A* with back-end control point optimization.

The results demonstrate that our proposed method achieves a significantly higher success rate compared to ego-planner. It is worth noting that the few successful trials of ego-planner mainly occurred when the person moved randomly and crossed the aerial robot’s planned route ahead of time, thereby avoiding direct interaction with the dynamic obstacle. This incidental effect also explains its shorter planning time. In contrast, ours (only R-A*) achieved a noticeably higher success rate than ego-planner, owing to the introduction of risk modeling during path search. However, the lack of back-end dynamic obstacle avoidance optimization resulted in a lower overall success rate compared to the ours(full) method.

In summary, the experiments validate the necessity of integrating risk-aware path search in the front end with high-risk dynamic obstacle avoidance optimization in the back end. The synergy of both modules significantly enhances planning robustness and success rate in complex dynamic environments.

\begin{figure}
    \centering
    \includegraphics[width=1.0\linewidth]{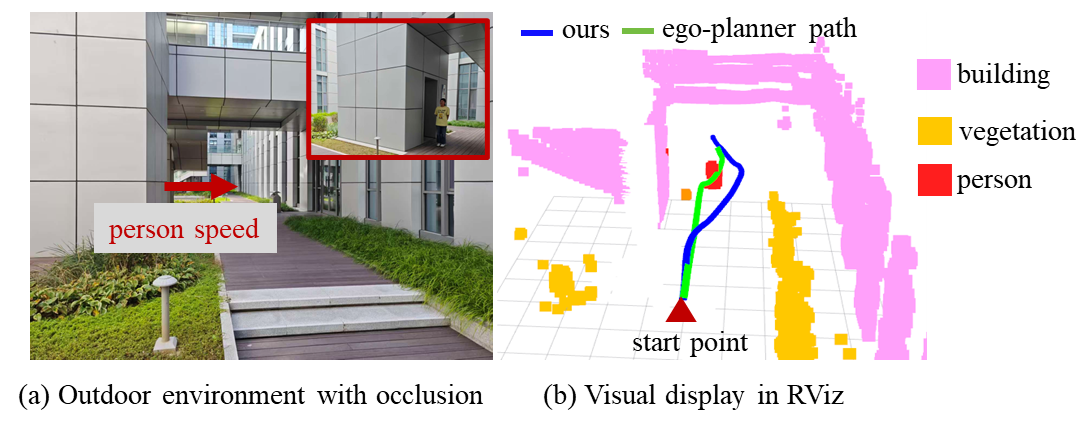}
    \caption{RA-Nav can predict potential risks in occluded environments based on semantic information and plan collision-free trajectories.}
    \label{fig:placeholder}
\end{figure}

\begin{table}[t]
\centering
\caption{Comparison of success rate, planning time, and flight time in two simulated scenes.}
\resizebox{\columnwidth}{!}{ 
\begin{tabular}{ccccc}
\toprule
Scene & Method & Success Rate & Planning Time & Flight Time \\
      &        & (\%)         & (ms)          & (s) \\
\midrule
\multirow{3}{*}{1} 
& Ego-planner \cite{c4}    & 25  & 0.113 & 3.50 \\
& Ours (only R-A*) & 70  & 0.447 & 4.29 \\
& Ours (full)         & 100 & 0.727 & 6.09 \\
\midrule
\multirow{3}{*}{2} 
& Ego-planner \cite{c4}    & 20  & 0.165 & 6.56 \\
& Ours (only R-A*) & 75  & 0.368 & 6.91 \\
& Ours (full)          & 95  & 0.532  & 7.73 \\
\bottomrule
\end{tabular}}
\label{tab:simulation_results}
\end{table}
To further validate the system’s practical applicability, we test it with point cloud data collected in real-world environments using the MID360 lidar. In the Fig.\ref{fig:placeholder}, a person emerges from behind a building as the drone passes by. The point clouds are processed through LMSCNet to obtain semantic segmentation results. In the outdoor scenario, RA-Nav proactively avoids occluded corners and generates collision-free trajectories, outperforming the ego-planner. This is achieved by constructing a risk grid map from semantic information, which serves as the planner’s local map input and enables the drone to avoid high-risk areas.

\section{Conclusions}
\label{conclusions}
A semantic map-based obstacle avoidance framework for aerial robots operating in high-risk environments is proposed in this paper. A multi-scale semantic segmentation module was employed to identify obstacle categories and assign predefined risk weights, enabling the construction of risk-aware grid maps. By modeling risk fields through Gaussian superposition and incorporating gradient information, the proposed method effectively handles suddenly moving obstacles. Extensive simulations demonstrate that RA-Nav significantly improves planning robustness. In particular, the success rate improves from 20$\%$ (baseline) to 95$\%$ (RA-Nav) in simulated unpredictable scenarios. Furthermore, its effectiveness is validated through simulations using real-world environmental data.



\end{document}